\documentclass{ifacconf}

\usepackage{graphicx}      
\usepackage{natbib}        
\usepackage{amsmath}
\usepackage{amssymb}
\usepackage{algorithm}
\usepackage{algpseudocode}
\usepackage{booktabs}
\usepackage{multirow}
\usepackage{xcolor}

\begin{document}
\begin{frontmatter}

{\color{red} \textbf{© 2023 the authors. This work has been accepted to IFAC for publication
under a Creative Commons Licence CC-BY-NC-ND}}

\title{Using Reinforcement Learning to Simplify Mealtime Insulin Dosing for People with Type 1 Diabetes: \textit{In-Silico} Experiments \thanksref{footnoteinfo}} 

\thanks[footnoteinfo]{Supported by American Diabetes Association grant \#4‐22‐PDFPM‐16.}

\author[First]{Anas El Fathi} 
\author[Second]{Marc D. Breton} 

\address[First]{University of Virginia, Center for Diabetes Technology, Charlottesville, VA, USA (e-mail: fwt9vd@virginia.edu).}
\address[Second]{University of Virginia, Center for Diabetes Technology, Charlottesville, VA, USA (e-mail: mb6nt@virginia.edu)}

\begin{abstract}                
People with type 1 diabetes (T1D) struggle to calculate the optimal insulin dose at mealtime, especially when under multiple daily injections (MDI) therapy. Effectively, they will not always perform rigorous and precise calculations, but occasionally, they might rely on intuition and previous experience. Reinforcement learning (RL) has shown outstanding results in outperforming humans on tasks requiring intuition and learning from experience. In this work, we propose an RL agent that recommends the optimal meal-accompanying insulin dose corresponding to a qualitative meal (QM) strategy that does not require precise carbohydrate counting (CC) (e.g., a usual meal at noon.). The agent is trained using the soft actor-critic approach and comprises long short-term memory (LSTM) neurons. For training, eighty virtual subjects (VS) of the FDA-accepted UVA/Padova T1D adult population were simulated using MDI therapy and QM strategy. For validation, the remaining twenty VS were examined in 26-week scenarios, including intra- and inter-day variabilities in glucose. \textit{In-silico} results showed that the proposed RL approach outperforms a baseline run-to-run approach and can replace the standard CC approach. Specifically, after 26 weeks, the time-in-range ($70-180$mg/dL) and time-in-hypoglycemia ($<70$mg/dL) were $73.1\pm11.6$\% and $ 2.0\pm 1.8$\% using the RL-optimized QM strategy compared to $70.6\pm14.8$\% and $ 1.5\pm 1.5$\% using CC. Such an approach can simplify diabetes treatment resulting in improved quality of life and glycemic outcomes.

\end{abstract}


\end{frontmatter}

\section{Introduction}

Type 1 diabetes (T1D) is an autoimmune disease treated by exogenous insulin to maintain blood glucose levels at clinically acceptable targets. Sustained elevated glucose levels (hyperglycemia) lead to long-term complications, including cardiovascular complications, kidney failure (neuropathy), and eye complications (retinopathy) (\cite{nathan1993effect}). Fear of low glucose levels (hypoglycemia) and its acute complications are major limiting factors to achieving glucose targets. The gold standard T1D treatment involves intensive insulin therapy, where basal and bolus insulin are delivered, and glucose levels are regularly monitored. Basal insulin aims to maintain glucose levels constant during fasting and overnight. Bolus insulin is given at mealtimes to compensate for the significant glucose increase due to meal carbohydrates. The most common approach to implement intensive insulin therapy is termed "Multiple Daily Injections" (MDI) and employs insulin syringes or pens.

The large inter-individual and intra-individual variability in insulin sensitivity requires individualization and continuous adaptation of mealtime insulin doses (\cite{heinemann2002variability}). Glucose control after meals is a challenging part of diabetes and a major contributor to overall degraded glycemic control (\cite{el2018artificial}). Since carbohydrates are the primary determinant of the postprandial response, patients are encouraged to perform carbohydrate counting (CC) to calculate a matching insulin bolus (\cite{sheard2004dietary}. CC involves the use of individualized parameters, such as the insulin-to-carbohydrate ratio (ICR), a ratio indicating the amount of carbohydrate covered by a unit of insulin, the correction factor (CF), a factor indicating the drop in glucose levels caused by a unit of insulin, and the glucose target (\cite{walsh2011guidelines}). In practice, proper CC requires rigorous education and good numeracy skills, is prone to human errors, requires continuous adaptation and optimization of therapy parameters (e.g., ICR, CF, glucose target), and is associated with increased disease management burden (\cite{roversi2020modeling, fortin2017practices, mannucci2005eating}). As a result, most people with T1D only estimate the amount of mealtime insulin bolus using their experience with previously consumed meals: they rely on empirical estimates and learn by trial and error from their previously consumed meals and delivered insulin doses.

Reinforcement learning (RL) is a framework for learning decision-making strategies by interacting with an environment. In this framework, an agent learns to map a situation (a state) to an action resulting in maximum cumulative future rewards (\cite{sutton2018reinforcement}). Deep neural networks (DNN) unleashed the full potential of the RL algorithm by serving as universal function approximations and eliminating the need to configure features manually. As noted in this recent review by \cite{tejedor2020reinforcement}, RL is increasingly being explored in the diabetes field to optimize insulin dosing strategies. Although such algorithms have not yet been used in clinical practice, a few clinical trials using RL have been planned (\cite{jafar2021long}, \cite{zhu2020basal}).

In this work, RL is explored as a possibility to replace human intuition in learning the optimal policy for mealtime insulin dosing using limited carbohydrate content information. The RL agent is trained, using only the historical glucose and insulin data, to optimize the insulin dose corresponding to  a qualitative meal (QM) strategy where meals are categorized instead of explicitly counted (e.g., a `snack' instead of an exact 12g of carbohydrate). Such an algorithm can be used in a novel insulin bolus calculator that does not require carbohydrate counting. Thus, simplifying insulin therapy and potentially improving quality of life and glycemic outcomes.  

\section{Methods}

\subsection{The Soft-Actor-Critic approach}

 The RL problem can be described as a Markov Decision Process (MDP) consisting of a set of states $\mathcal{S}$, a set of actions $\mathcal{A}$, a transition distribution $\mathcal{D}$, a reward function $R(.)$, and a discount factor $\gamma < 1$. The objective of an RL agent is to optimize the future \textit{return}: a value corresponding to the summation of discounted future rewards. In an actor-critic paradigm, the critic agent learns to evaluate the usefulness of action $a$ at a state $s$ in terms of optimal returns. The actor agent uses the critic evaluation to search for the best actions. Explicitly, the critic learns the action-value function defined in (\ref{eq:actionvalue}).
\begin{equation} \label{eq:actionvalue}
    Q^{\pi}(s, a) = \mathop{\mathbb{E}}_{\substack{s^\prime \sim \mathcal{D} \\ a^\prime \sim \pi}}\left[ \sum_{k=t}^{\infty} \gamma^{k-t+1} R_k | s_t=s, a_t=a \right]
\end{equation}

The Soft Actor-Critic (SAC) algorithm optimizes a stochastic policy in an off-policy way (learning can happen from a replay buffer with a history of actions) while considering an entropy term measuring randomness in the policy (\cite{haarnoja2018soft}). The policy is therefore trained to maximize a trade-off between expected returns and exploring new actions. In this context, the SAC recursive Bellman equation is defined as follows:

\begin{equation} \label{eq:bellmanEqn}
Q^{\pi}(s, a) = \mathop{\mathbb{E}}_{\substack{s^\prime \sim \mathcal{D} \\ a^\prime \sim \pi}}\left[ R_t + \gamma \left(Q^{\pi}(s^\prime, a^\prime) + \alpha \log{\pi(a^\prime | s^\prime)} \right)  \right]
\end{equation}

where $a^\prime \sim \pi(., s^{\prime})$ and $\alpha$ is the entropy temperature parameter. This parameter is also trained to match the desired target entropy $\mathcal{H}$: 

\begin{equation} \label{eq:entropyMax}
    \min_{\alpha_t} \left(-\alpha_t \left( \log{\pi(a^\prime | s^\prime)} + \mathcal{H} \right) \right)
\end{equation}

Due to instabilities and evaluation over-estimation, it is recommended to use multiple (in this case, two) critic networks and corresponding target critic networks that are slowly updated to sync with the original networks. In brief, the weight $\phi$ of the critic networks are trained to minimize the Euclidean distance to the target $y(r, s^\prime)$ defined as follows:

\begin{equation} \label{eq:bellmanTarget}
y(r, s^\prime) = r + \gamma \left(\min_{i} Q_{\phi_{\text{targ}_i}}^{\pi}(s^\prime, a^\prime) + \alpha \log{\pi(a^\prime | s^\prime)} \right)
\end{equation}

Because we are dealing with a stochastic continuous action, a reparameterization is performed to map a random variable $\epsilon \sim \mathbb{N}(0, 1)$ to action as follows:

\begin{equation} \label{eq:policyForm}
a_{\theta}(s, \epsilon) = a_{\text{max}} \tanh{\left(\mu_\theta(s) + \sigma_\theta(s) \cdot \epsilon\right)} \quad \epsilon \sim \mathbb{N}(0, 1)
\end{equation}

The weights of $\mu_\theta(s)$ and $\sigma_\theta(s)$ can be learned by maximizing the critic evaluation:

\begin{equation} \label{eq:policyMax}
\max_{\theta}{\mathop{\mathbb{E}}_{\substack{s^\prime \sim \mathcal{D} \\ \epsilon \sim \mathbb{N}}}\left[ \min_{i} Q_{\phi_{   i}}^{\pi}(s, a) - \alpha \log{\pi(a_{\theta}(s, \epsilon) | s)} \right]}
\end{equation}

The final Soft-Actor-Critic approach is in Algorithm \ref{alg:cap}:

\begin{algorithm}
\caption{Soft Actor-Critic for T1D simulator}\label{alg:cap}
\begin{algorithmic}[1]
\Require T1D Simulation environment with 80 VS for training and 20 VS for validation 
\While{$R_{val}$ increases}
\State Observe $s$, select and execute action $a \sim \pi(.|s)$
\State Observe $s^\prime$, $r$ and indicate if episode ended $d$
\State Store $\left(s ,\ a ,\ r ,\ s^\prime ,\ d\right)$ in replay buffer $\mathcal{R}$
\If{d}
Reset environment
\EndIf
\While{$\left(s ,\ a ,\ r ,\ s^\prime ,\ d\right)$ in training batches}
\State Optimize critic networks by ``(\ref{eq:bellmanTarget})''.
\State Optimize actor network by ``(\ref{eq:policyMax})''.
\State Optimize entropy temperature by ``(\ref{eq:entropyMax})''. 
\State Update Target networks by 
\begin{equation}
    \phi_{\text{targ}_i} \leftarrow \rho \phi_{\text{targ}_i} + (1 - \rho) \phi_{i}
\end{equation}
\EndWhile
\State Run validation and update $R_{val}=\sum_{i=0}^{i=T}\gamma^{T-i} r(s_i, a_i, s_{i+1})$
\EndWhile
\end{algorithmic}
\end{algorithm}

\subsection{Problem formulation}

To construct the qualitative mealtime (QM) insulin dosing strategy, we propose a new formulation of meal-accompanying insulin bolus, not requiring carbohydrate content but using a meal category. The mealtime insulin dose is calculated, as shown in (\ref{eq:bolusCalc}), using a qualitative meal description $c_i$, the time of the day $t_j$, the current glucose level $G$, the glucose target $G_t$, the insulin sensitivity factor $K_{ISF}$, and the amount of insulin that is still not absorbed into the blood circulation: insulin-on-board (IOB).

\begin{equation} \label{eq:bolusCalc}
B(c_i, t_j) = b_{c_i} \alpha_{t_j} + \max\left(\frac{G - G_t}{K_{ISF}}, 0\right) - IOB
\end{equation}

where $b_{c_i}$ is the insulin bolus for meal-category $c_i$, and $\alpha_{t_j}$ is a coefficient depending on the time of the insulin bolus $t_j$. We define four QM categories: a snack, a less-than-usual meal, a usual meal, and a more-than-usual meal, and six time intervals of four hours each (3am$-$7am, 7am$-$11am, 11am$-$3pm, 3pm$-$7pm, 7pm$-$11pm, 11pm$-$3am). With this formulation, the objective is to optimize $(b_{c_i})_{i\in \{1\dots4\}}$ and $(\alpha_{t_j})_{j\in \{1\dots6\}}$ using historical glucose and insulin data.

\subsection{Proposed reinforcement learning agent}

\subsubsection{Neural Network Architecture}

In deep reinforcement learning (DRL), the  actor and critic are parameterized as DNNs with trainable weights. Because of the sequential nature of our observation (glucose time series, meal categories time series), we employ a long short-term memory (LSTM) architecture. LSTMs are a form of recurrent neural network where a feedback connection enables processing sequences while accounting for lags of unknown duration between important events in the time series (\cite{hochreiter1997long}). 

\begin{figure}
\begin{center}
\includegraphics[width=8.86cm]{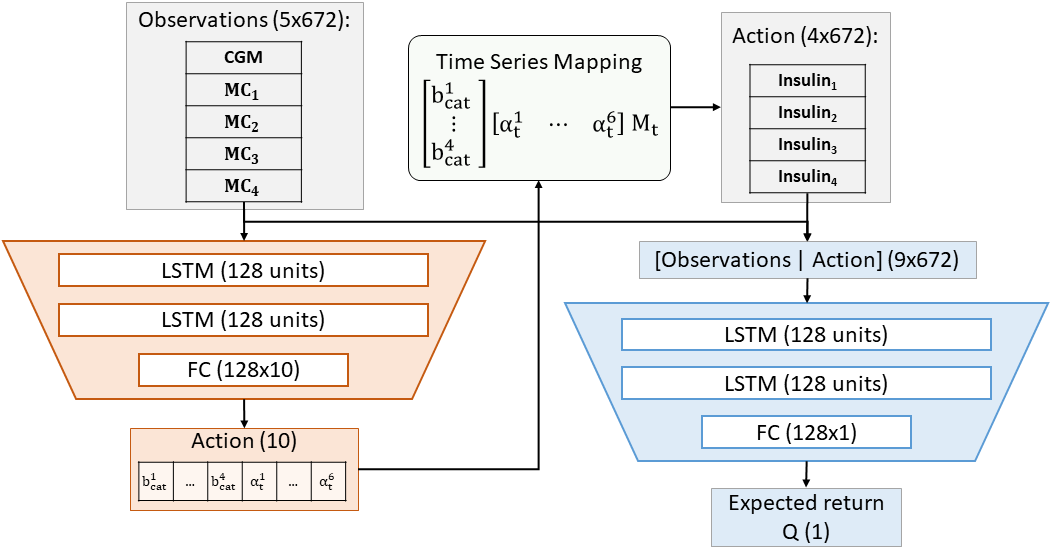}    
\caption{Architecture for the actor-critic DNN. $(MC_i)_{1:4}$ are meal event indicators. $(Insulin_i )_{1:4}$ is the insulin time series for each meal category. $M_t$ is a matrix that maps each action to the time when it is applied. FC stands for a fully-connected  network. Dimensions are noted between parentheses.} 
\label{fig:architecture}
\end{center}
\end{figure}

Figure \ref{fig:architecture} shows a diagram of the actor and critic networks. The actor inputs are the observation time series and output an encoded action $A(s)$. The critic network combines the action and state to produce an expected return $Q(s, a)$ 

\subsubsection{Actions}

The agent outputs ten separate actions: four actions representing the change in the insulin dose for each category $\delta b_{c_i}$ and six actions representing the change in the insulin dose, regardless of the meal category, for a specified 4-hour window of the day $\delta \alpha_{t_j}$. 

\begin{equation} \label{eqn:action}
    a_{\theta} \sim \left\{\begin{bmatrix}
    \delta b_{c_1}  & \dots  & \delta b_{c_4} & \delta \alpha_{t_1}  & \dots  & \delta \alpha_{t_6}
    \end{bmatrix}\right | a_{{\theta}_i} \in [-1, 1]\}
\end{equation}

The 10-dimensional action can be transformed into four time series where each corresponds to a meal category and indicates when the action will be taken. As presented in Fig. \ref{fig:architecture}, this transformation uses a matrix ($M_{Time}$) where the row corresponds to when $\delta \alpha_{t_j}$ is applied.

\subsubsection{States}

The inputs are formed as a multi-dimensional vector informing the state of the user from collected observations. Following the international consensus on continuous glucose monitoring (CGM) metrics by \cite{battelino2019clinical}, we choose to observe 14 days of glucose data. Collected CGM is sampled with a 30-minute step, resulting in a time series of length 672. Because of the asymmetric relevance of low glucose values compared to high glucose values, we normalize the CGM by a log-transform and a re-scaling between $\left[-1, 1\right]$ as shown in (\ref{eqn:cgmnorm}).

\begin{equation} \label{eqn:cgmnorm}
    G_{norm} = \dfrac{2 \log(G) - (\log(G_{max}) + \log(G_{in}))}{\log(G_{max}) - \log(G_{in})}
\end{equation}

where $G_{max}=400$ and $G_{min}=40$.

Because of delays in insulin absorption, actions only affect glucose after an important temporal lag. To enforce this, the actions are encoded for each meal category by fractions of $1$ when insulin is still active, i.e., "1" when the meal occurs, followed by fractions of "1" to match the insulin absorption profile. 

\subsubsection{Rewards}

the reward is a proxy for the desired objective function. Using the consensus CGM glucose targets in \cite{battelino2019clinical}, we define the following reward:

\begin{align} \label{eq:reward}
r_{hypo}( s^\prime) = 
& \max{\left(\frac{T_{<54}^{s^\prime}}{T_{<54}^{th}} - 1,\ 0\right)} + \max{\left(\frac{T_{<70}^{s^\prime}}{T_{<70}^{th}} - 1,\ 0\right)} \nonumber \\
r_{hyper}( s^\prime) = 
& \max{\left(\frac{T_{>180}^{s^\prime}}{T_{>180}^{th}} - 1,\ 0\right)} + \max{\left(\frac{T_{>250}^{s^\prime}}{T_{>250}^{th}} - 1,\ 0\right)}  \nonumber \\
r^{-}(s^\prime) = & r_{hypo}( s^\prime) + r_{hyper}( s^\prime) \nonumber \\
r^{+}(s^\prime) = & \max{\left(\frac{T_{\geq70 \ \leq180}^{s^\prime}}{1 - T_{>180}^{th} - T_{>70}^{th}} - 1,\ 0\right)} \nonumber \\
r(s^\prime) = & \begin{cases}
-r^{-}(s^\prime) & \text{if} \quad r^{-}(s^\prime) > 0 \\ 
r^{+}(s^\prime) & \text{otherwise}
                \end{cases}
\end{align}

where $T^{s}_{X}$ is the percentage of time spent in the interval $X$ (e.g., $T^{s}_{<54}$ indicates less than 54mg/dL) while observing the state $s$, and $T^{th}_{X}$ are thresholds set as follows: $T_{<54}^{th} = 1\%$, $T_{<70}^{th} = 4\%$, $T_{>180}^{th} = 25\%$, $T_{>250}^{th} = 5\%$.

\subsubsection{Training}

Multiple strategies were employed to accelerate and stabilize the training.  Dropouts (the process of randomly dropping out NN units with a predefined probability) were used between layers. A "prioritized replay experience" replay buffer was used where transitions are sampled with a probability proportional to their importance priority. The "importance priority" is updated for each base using the critic network error (\ref{eq:bellmanTarget}). The training was executed for $2,500$ epochs starting from different seeds. A replay ratio of 1.0 was used (Table \ref{tb:hyperparam}), meaning that in each epoch, we generate one transition per virtual subject (VS) (80 transitions) and train in the replay memory for $80\times1=80$ episodes where each episode has a size of batch size (256). In total, $200,000$ simulations were performed.

\begin{table}[hb]
\begin{center}
\caption{Training hyper-parameters}\label{tb:hyperparam}
\begin{tabular}{cc}
\toprule
\textbf{Parameter} & \textbf{Value} \\
\midrule
Replay buffer capacity & $10^6$ \\
Replay ratio & 1.0 \\
Target smoothing coefficient & $0.005$ \\ 
Critic-network learning rate & $4\times10^{-4}$ \\ 
Actor-network learning rate & $2\times10^{-4}$ \\ 
Entropy learning rate & $4\times10^{-4}$ \\ 
Batch size & 256 \\
Number of epochs &  $2500$ \\
\bottomrule
\end{tabular}
\end{center}
\end{table}

\subsection{Baseline algorithm for simplified mealtime insulin}


As a baseline, we propose a run-to-run (R2R) optimization algorithm that alters the bolus dose in each run to optimize postprandial glucose exposure. We denote $\delta b_{c_i, t_j}$ the change in the bolus corresponding to a meal category $c_i$ and a time $t_j$. The change in this dose $\delta b_{c_i, t_j}$ relates to the action changes $\delta b_{c_i}$ and $\delta \alpha_{t_j}$, as 

\begin{equation} \label{eq:diffBolus}
\delta b_{c_i, t_j} = \delta b_{c_i} + \delta \alpha_{t_j} + \delta b_{c_i} \times \delta \alpha_{t_j}
\end{equation}

and it can be updated using the R2R rule:

\begin{equation} \label{eq:run2run}
\delta b_{c_i, t_j} = \begin{cases} 
                 - k_{hypo} r_{hypo}( b_{c_i, t_j}) & \text{if} \quad r_{hypo}( b_{c_i, t_j}) > 0 \\ 
                 k_{hyper} r_{hyper}( b_{c_i, t_j}) & \text{if} \quad r_{hyper}( b_{c_i, t_j}) > 0 \\
                0 & \text{otherwise}
              \end{cases} 
\end{equation}

where $r_{hypo}$ and $r_{hyper}$ are defined in the reward function (\ref{eq:reward}), and $k_{hypo}$, $k_{hyper}$ are gains set to $0.1$. Once $\delta b_{c_i, t_j}$ is computed, the action ($\delta b_{c_i}$ and $\delta \alpha_{t_j}$) can be solved using (\ref{eq:diffBolus}) and an additional regularization term.

\subsection{Experimental setup}

The adult cohort of the FDA-accepted UVA/Padova T1D simulator is used to run the \textit{in-silico} experiments. 80 VS were used during the training phase, and 20 VS were kept for validation. Each step consists of a 14-day simulation of MDI therapy following the QM dosing strategy.
Specifically, each VS consumes randomly generated meals with different sizes at distinct times. For each meal, the VS quantifies the meal to one of the four categories using three randomly selected thresholds (snacks are between $0$ and $m_1 \sim \mathcal{U}(10, 50)$, less than usual meals are between $m_1$ and $m_2 \sim \mathcal{U}(60, 90)$, usual meals are between $m_2$ and $m_3 \sim \mathcal{U}(100, 120)$, are more than usual meals are greater than $m_3$). At the beginning of each episode, the VS is initialized with a randomly generated QM strategy (($b_{c_i}$ and $\alpha_{t_j}$)). 
During one episode, in each step, new actions ($\delta b_{c_i}$, $\delta \alpha_{t_j}$) are produced by the agent (RL or R2R) and are used in the following step (14-day simulation). An episode ends if any VS has a negative glucose measurement (over-delivering insulin) or a maximum step of 64 is achieved.

During training, an idealized scenario is used where consumed meals are the same during the episode (each episode has a random meal plan), and there is no variability in the VS model parameters. For validation, day-to-day variability is modeled by daily varying the parameters $k_{p_1}$ and $Vm_0$ and randomly selecting the time and size of meals in each step (see \cite{dalla2007meal} for the complete set of model equations). All \textit{in-silico} experiments were performed in a C++ implementation of the UVA/Padova simulator to accelerate the runtime.

In order to evaluate the performance of the proposed approach, two \textit{in-silico} experiments are carried out. In both experiments, for 26 weeks, 20 virtual subjects are simulated using: (i) (\textbf{CC}) an informed carbohydrate counting strategy where VS can make errors ($\pm20\%$) in carbohydrate counting as documented by \cite{brazeau2013carbohydrate}; (ii) (\textbf{QM-Default}) a meal categorization strategy where insulin bolus strategy is not optimized and fixed during the 26 weeks; (iii) (\textbf{QM-R2R}) a meal categorization strategy where insulin bolus strategy is optimized using the baseline R2R algorithm; (iv) (\textbf{QM-RL}) a meal categorization strategy where insulin bolus strategy is optimized using the proposed RL algorithm. In one experiment, daily random meals and day-to-day variability are used (\textbf{Var}), and in the other, there is no variability, and the same meal plan is given throughout (\textbf{Simple}). The \textbf{Simple} scenario is similar to the training scenarios.

\subsection{Outcomes metrics}

Algorithm performance was assessed using established metrics of glycemic variability and quality of glycemic control (\cite{battelino2019clinical}), including time $<54$ mg/dL (TBR2); time $<70$ mg/dL (TBR1); time in the target $70-180$ mg/dL (TIR); time $>180$ mg/dL (TAR1); time $>250$ mg/dL (TAR2); Mean glucose (mg/dL); standard deviation (SD) of glucose (mg/dL); and total bolus insulin (U). Metrics were calculated every 14 days. 

\section{Results}

\begin{figure}
\begin{center}
\includegraphics[width=8.8cm]{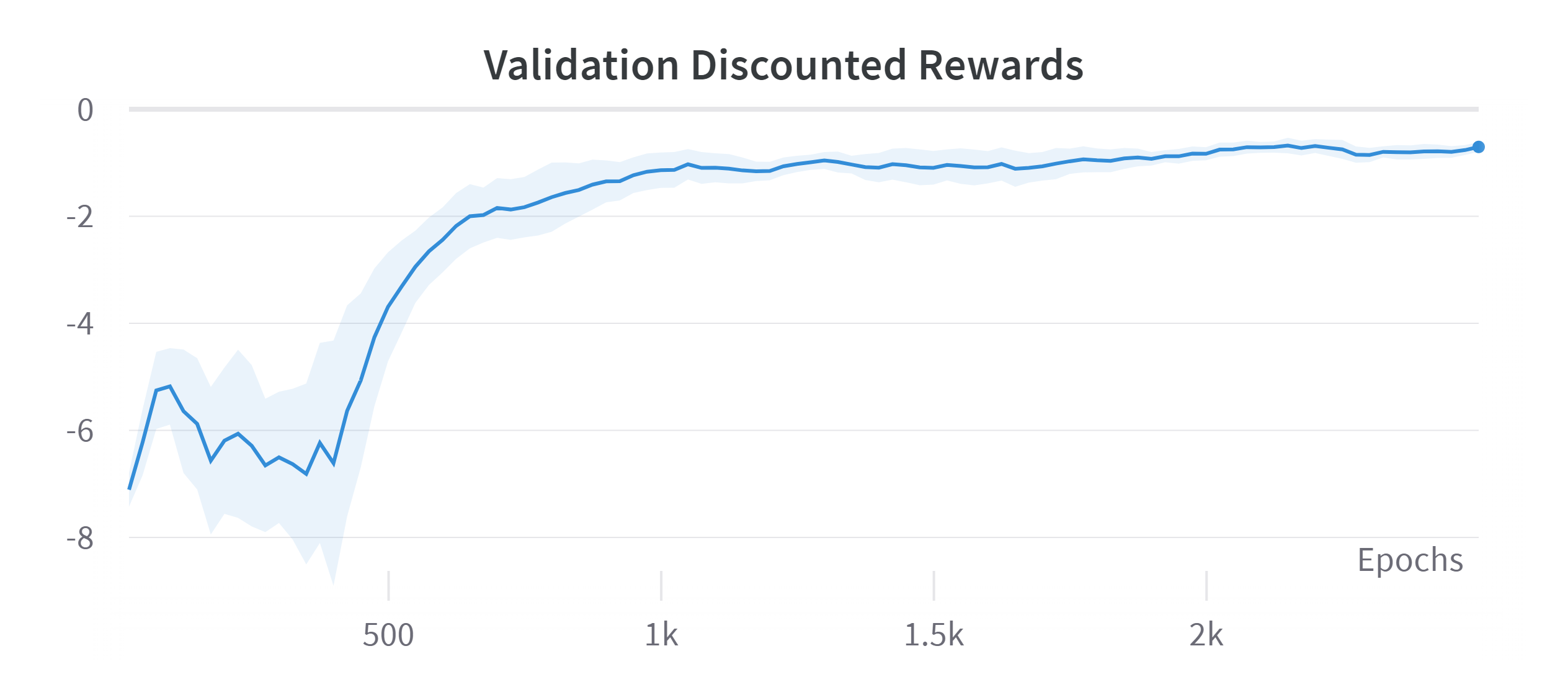}    
\caption{Total discounted rewards in the validation set during training: 20 virtual subjects simulated for 26 weeks using the \textbf{Var} configuration. The graph shows the mean and standard error of three training runs.} 
\label{fig:validation}
\end{center}
\end{figure}

\begin{figure}
\begin{center}
\includegraphics[width=8.8cm]{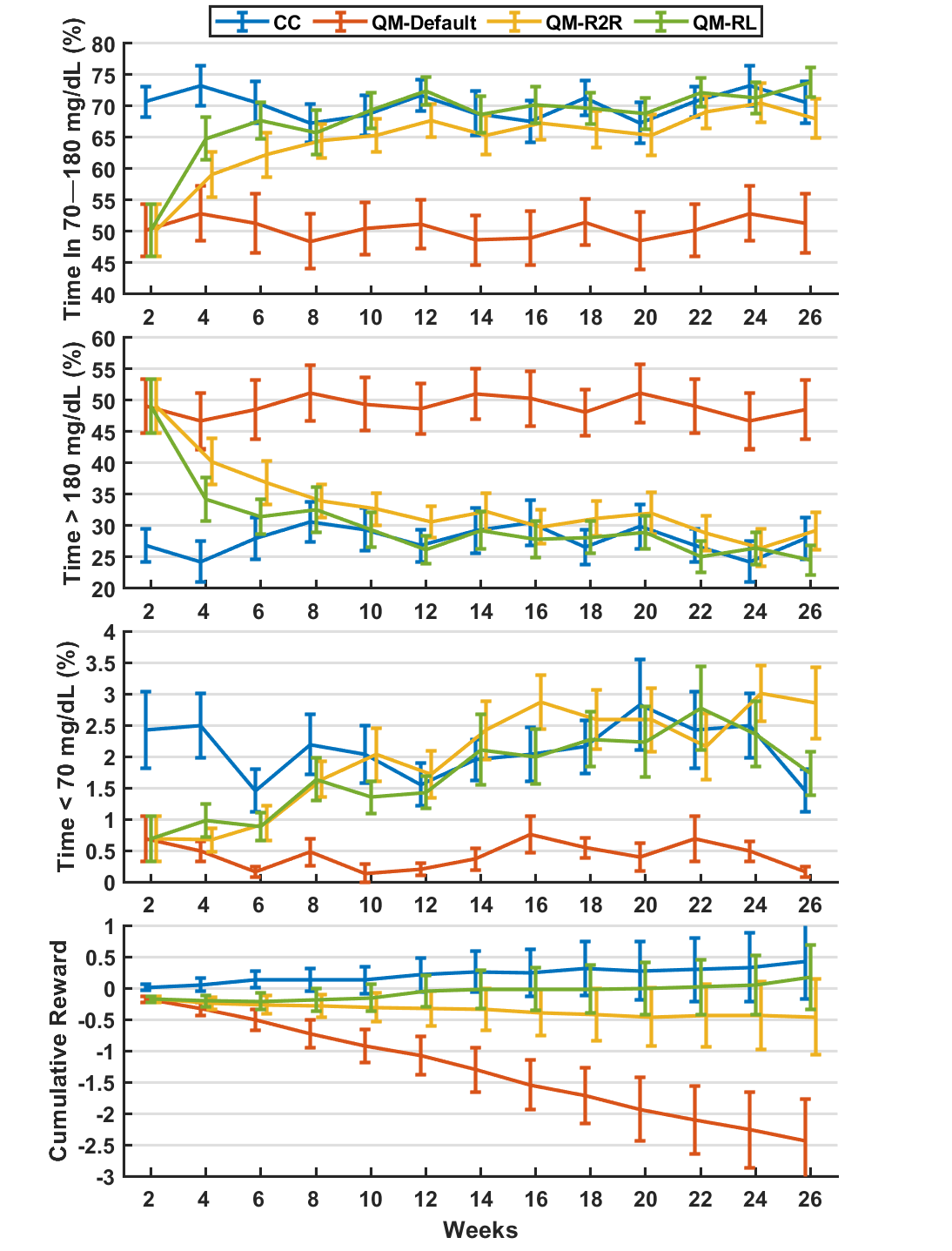}    
\caption{Summary of 26-weeks \textit{in-silico} experiment in the \textbf{Var} scenario. Values are represented as mean and standard errors for each two weeks period.} 
\label{fig:val_var_weeks}
\end{center}
\end{figure}

\begin{figure*}
\begin{center}
\includegraphics[width=18cm]{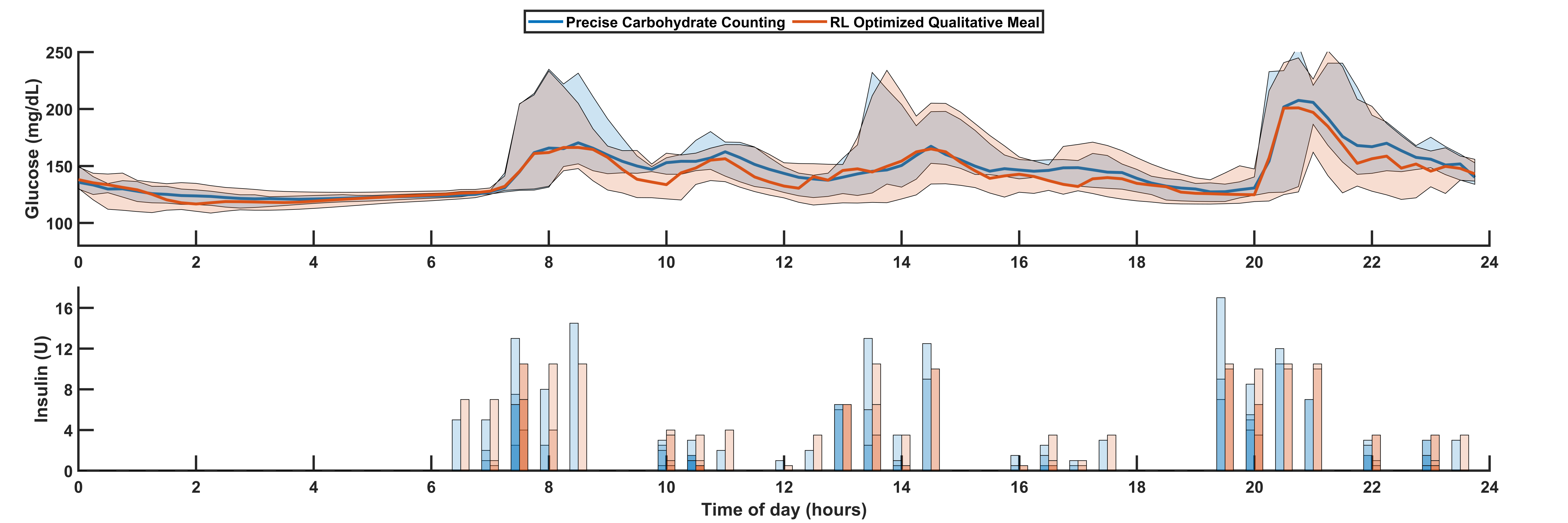}    
\caption{Top: Summary glucose profile of the last two weeks of the \textbf{Simple} \textit{in-silico} experiment. The envelope is the median and interquartile range. Bottom: Summary of all insulin doses given during the last two weeks of the \textbf{Simple} \textit{in-silico} experiment. Darker color indicates multiple insulin doses.} 
\label{fig:example}
\end{center}
\end{figure*}

Fig. \ref{fig:validation} shows the total reward in the validation virtual subjects using different seeds by training epoch. Running the $2,500$ epochs takes 80 hours on a 64-core, 128 GB RAM Linux machine with an NVIDIA GeForce RTX 2080 Ti 11 GB GPU with 4352 cores. The runtime was limited by the time needed to simulate the transitions, where it takes $\sim15s$ to simulate $80$ virtual subject for a 14-day period. Table \ref{tab:outcomes} summarizes the outcomes in two variability scenarios. In the \textbf{Simple} scenario, after 26 weeks, the TIR and TBR were $81.1\pm9.5$\% and $ 0.7\pm 1.4$\%, respectively, using RL-optimized QM strategy compared to $82.9\pm12.6$\% and $ 0.4\pm 1.0$\% using precise carbohydrate counting and ideal therapy parameters. In the \textbf{Var} scenario, after 26 weeks, the time-in-range TIR and TBR were $73.1\pm11.6$\% and $ 2.0\pm 1.8$\%, respectively, using RL-optimized QM strategy compared to $70.6\pm14.8$\% and $ 1.5\pm 1.5$\% using imprecise carbohydrate counting and ideal therapy parameters. In both scenarios, the RL approach outperforms the baseline run-to-run algorithm. 
Fig. \ref{fig:val_var_weeks} shows the evolution of the outcomes over the 26-week simulation in the \textbf{Simple} scenario. Fig. \ref{fig:example} shows a summary of the last two weeks of glucose and insulin data for one of the VS in the \textbf{Var} scenario. 

\begin{table*}[t]
\setlength\tabcolsep{3.8pt}
\begin{center}
\caption{Summary of outcomes of two \textit{in-silico} experiments comparing CC and qualitative meal strategy in two scenarios (\textbf{Var} vs. \textbf{Simple}). Values are Mean $\pm$ standard deviation.}
\label{tab:outcomes}
\begin{tabular}{l|c|ccc|c|ccc}
\toprule
& \multicolumn{4}{c|}{\textbf{Simple}} & \multicolumn{4}{c}{\textbf{Var}}  \\
\cline{2-5}
\cline{6-9}
& \textbf{Precise} & \multicolumn{3}{c|}{\textbf{Qualitative Meal}} & \textbf{Imprecise} & \multicolumn{3}{c}{\textbf{Qualitative Meal}} \\
\cline{3-5}
\cline{7-9}
& \textbf{CC} & \textbf{Default} & \textbf{R2R} & \textbf{RL} & \textbf{CC} & \textbf{Default} & \textbf{R2R} & \textbf{RL} \\
\midrule
\textbf{TBR2 (\%)}  & $ 0.1\pm 0.2$ & $\pmb{ 0.0\pm 0.0}$ & $ 0.9\pm 1.1$ & $\mathit{ 0.3\pm 0.8}$ & $ 0.6\pm 0.8$ & $\pmb{ 0.0\pm 0.1}$ & $ 1.4\pm 1.5$ & $\mathit{ 0.5\pm 0.6}$\\ 
\textbf{TBR1 (\%)}  & $ 0.4\pm 1.0$ & $\pmb{ 0.0\pm 0.0}$ & $ 2.1\pm 1.5$ & $\mathit{ 0.7\pm 1.4}$ & $ 1.5\pm 1.5$ & $\pmb{ 0.2\pm 0.4}$ & $ 2.9\pm 2.6$ & $\mathit{ 1.7\pm 1.5}$\\ 
\textbf{TIR (\%)}  & $82.9\pm12.6$ & $53.6\pm19.5$ & $\mathit{74.4\pm11.6}$ & $\pmb{81.1\pm 9.5}$ & $70.6\pm14.8$ & $51.3\pm21.2$ & $\mathit{68.0\pm14.0}$ & $\pmb{73.8\pm10.8}$\\ 
\textbf{TAR1 (\%)}  & $16.7\pm12.7$ & $46.4\pm19.5$ & $\mathit{23.5\pm10.2}$ & $\pmb{18.2\pm 9.4}$ & $27.9\pm15.0$ & $48.5\pm21.3$ & $\mathit{29.1\pm13.5}$ & $\pmb{24.4\pm10.7}$\\ 
\textbf{TAR2 (\%)}  & $ 3.9\pm 5.7$ & $12.7\pm 9.2$ & $\mathit{ 5.1\pm 5.9}$ & $\pmb{ 4.7\pm 4.5}$ & $ 6.2\pm 6.1$ & $16.9\pm11.9$ & $\mathit{ 6.4\pm 5.6}$ & $\pmb{ 5.3\pm 4.0}$\\ 
\textbf{Mean (mg/dL)}  & $151.5\pm19.0$ & $189.2\pm25.6$ & $\mathit{155.6\pm15.1}$ & $\pmb{151.9\pm15.2}$ & $156.8\pm20.1$ & $189.0\pm30.2$ & $\mathit{157.6\pm18.2}$ & $\pmb{155.8\pm14.0}$\\ 
\textbf{SD (mg/dL)}  & $37.8\pm16.2$ & $53.6\pm18.5$ & $\mathit{48.5\pm17.2}$ & $\pmb{44.9\pm14.7}$ & $50.3\pm13.6$ & $61.8\pm18.7$ & $\mathit{53.9\pm15.6}$ & $\pmb{50.8\pm14.9}$\\ 
\midrule 
\textbf{Bolus (U)}  & $33.6\pm15.2$ & $23.4\pm11.5$ & $34.8\pm16.4$ & $33.6\pm16.1$ & $33.7\pm18.4$ & $23.9\pm13.8$ & $35.4\pm20.1$ & $32.0\pm15.5$\\ 
\bottomrule
\end{tabular}
\end{center}
\end{table*}

\section{Discussion}

To our knowledge, this is the first time optimization of a qualitative meal insulin bolus strategy that does not require carbohydrate counting has been proposed in MDI therapy. Such an approach has the potential to simplify treatment, resulting in improved quality of life and potentially improved glycemic outcomes. 

The proposed neural network employs an LSTM architecture that was able to encode the observed time series resulting in successful training. A simpler fully-connected architecture would have suffered from the curse of dimensionality where, without the LSTM weight-sharing, a larger network is needed. Because of the periodicity and sparsity of the signal, the use of attention mechanisms is another venue that can be explored in further research. 

Crafting reward functions for RL models can facilitate training. Originality we have used reward functions reported in the literature without being able to obtain similar results (\cite{zhu2020basal}, \cite{nordhaug2020silico}). Contrary to other work, the presented reward function intuitively reflects the accepted clinical targets. The proposed baseline R2R algorithm can also converge to satisfactory results by following an update rule derived from the same reward function.

The \textbf{Simple} experiment (Table \ref{tab:outcomes} and Fig. \ref{fig:example}) shows that the learned policy can achieve comparable results to the \textbf{Precise} scenario, meaning that it can compensate for non-optimal therapy parameters with optimized $b_{c_i}$ and $\alpha_{t_j}$ coefficients. Although the population outcomes are maintained, this strategy will never replace CC for an individual who is diligent with their CC and has optimized therapy parameters. The \textbf{Var} experiment is a challenging one (as seen in Fig. \ref{fig:val_var_weeks}, the outcomes in \textbf{QM-Default} and \textbf{CC} vary between weeks). Regardless, the learned policy is able to optimize the insulin bolus even though it was trained in simpler scenarios, not including random glucose and insulin fluxes. Furthermore, there was no explicit indication of the step size that should be taken in each step; however, the actor learned to produce big and precise steps when glucose is degraded and only make small stabilizing steps when glucose is on target. The proposed algorithm can be further improved by training on specific user data (domain adaptation), which was not explored.

\section{Conclusion}
In this work, an RL approach to optimize insulin doses in a novel mealtime dosing paradigm not requiring CC for people with T1D using MDI therapy is presented. \textit{In-silico} experiments demonstrate the feasibility of learning from historical glucose and insulin data, and the RL approach surpasses a baseline R2R algorithm using the same reward function. A mealtime bolus calculator using this algorithm has the potential to alleviate the need for carbohydrate counting, thus simplifying the therapy and improving glycemic outcomes. These results remain preliminary and limited to the conducted \textit{in-silico} experiments. Further research in RL for optimizing insulin dosing is warranted.


\bibliography{ifacconf}             


\end{document}